# Counterfactual thinking in cooperation dynamics

Luís Moniz Pereira[1] and Francisco C. Santos[2,3]

[1]NOVA-LINCS and Faculdade de Ciências e Tecnologia, Universidade Nova de Lisboa, Portugal

[2]INESC-ID and Instituto Superior Técnico, Universidade de Lisboa, Portugal

[3]ATP-Group, IST-Taguspark, Porto Salvo, Portugal

**Abstract.** Counterfactual Thinking is a human cognitive ability studied in a wide variety of domains. It captures the process of reasoning about a past event that did not occur, namely what would have happened had this event occurred, or, otherwise, to reason about an event that did occur but what would ensue had it not. Given the wide cognitive empowerment of counterfactual reasoning in the human individual, the question arises of how the presence of individuals with this capability may improve cooperation in populations of self-regarding individuals. Here we propose a mathematical model, grounded on Evolutionary Game Theory, to examine the population dynamics emerging from the interplay between counterfactual thinking and social learning (i.e., individuals that learn from the actions and success of others) whenever the individuals in the population face a collective dilemma. Our results suggest that counterfactual reasoning fosters coordination in collective action problems occurring in large populations, and has a limited impact on cooperation dilemmas in which coordination is not required. Moreover, we show that a small prevalence of individuals resorting to counterfactual thinking is enough to nudge an entire population towards highly cooperative standards.

## 1. Introduction

Counterfactual Thinking (CT) is a human cognitive ability studied in a wide variety of domains, namely Psychology, Causality, Justice, Morality, Political History, Literature, Philosophy, Logic, and AI [1-8]. CT captures the process of reasoning about a past event that did not occur, namely what would have happened had the event occurred, which may take into account what we know today. CT is also used to reason about an event that did occur, concerning what would have followed if it had not; or if another event might have happened in its place. An example situation: Lightning hits a forest and a devastating forest fire breaks out. The forest was dry after a long hot summer and many acres were destroyed. A counterfactual thought is: If only there had not been lightning, then the forest fire would not have occurred.

Given the wide cognitive empowerment of CT in the human individual, the question arises of how the presence of individuals with CT-enabled strategies affects the evolution of cooperation in a population comprising individuals of diverse interaction strategies. The natural locus to examine this issue is Evolutionary Game Theory (EGT) [9], given the amount of extant knowledge concerning different types of games, strategies, and techniques for the evolutionary characterization of such populations of individual players. EGT represents a dynamical and population-based counterpart of classical game theory. Having originated in evolutionary biology, EGT also holds a great potential in the realm of social sciences, given that human decision-making is often concerned within large populations and networks of self-regarding individuals.

The framework of EGT has been recently used to identify several of the key principles of mechanisms underlying the evolution of cooperation in living systems [9-11]. Most of these principles have been studied within the framework of two-person dilemmas. In this context, the Prisoner's Dilemma (PD) metaphor is possibly the most ubiquitously known game of cooperation. The dilemma is far more general than the associated story, and can be easily illustrated as follows. Two individuals have to decide simultaneously to offer (to *Cooperate*) or not (to *Defect*) a benefit $b$ to the other at a personal cost $c<b$. From a game theoretical point of view a ra-tional individual in a PD is always better off by not cooperating (defecting), irrespectively of the choice of the opponent, while in real life one often observes the opposite, to a significant extent. This apparent mismatch between theory and empirical results can be understood if one assumes an individual preference to cooperate with relatives, direct reciprocity, reputation, social structure, direct positive and negative incentives, among other sorts of community enforcing mechanisms (for an overview, see, e.g., [10, 11]).

Despite the popularity of the PD, other game metaphors can be used to unveil the mysteries of cooperation. Each game defines a different metaphor, the relative success of a player and attending strategy, and the ensuing behavioural dynamics. Moreover, while 2-person games represent a convenient way to formalize a pairwise cooperation, many real-life situations are associated with dilemmas grounded on decisions made by groups of more than 2 agents. Indeed, from group hunting, to modern collective endeavours such as *Wikipedia* and open source projects, or global collective dilemmas such as the management of common pool resources or the mitigation of the dangerous effects of climate change, general N-person problems are recurrent in biological and social settings.

The prototypical example of this situation is the N-person Prisoner's dilemma, also known as Public Goods Game. Here, $N$ individuals decide to contribute (or not) to a *public good*. The sum of all contributions is invested, and the returns of the investment shared equally among all group members, irrespectively of who contributed. Often, in these cases, free riding allows one to enjoy the public good at no cost, to the extent that others continue to contribute. If all players adopt the same reasoning, we are led to the *tragedy of the commons* [12], characterizing the situation in which everyone defects, with the making of cooperation a mirage. Below we will return to the details associated with the formalization of such dilemmas.

Importantly, depending on the game and associated strategies, individuals may revise their strategies in different ways. The common assumption of classic game theory is that players are rational, and that the Nash Equilibrium constitutes a reasonable prediction of what self-regarding rational agents adopt [13]. Often, however, players have limited cognitive skills or resort to simpler heuristics to revise their choices. Evolutionary game theory (EGT) [14] offers an answer to this situation, adopting a population description of game interactions in which individuals resort to social learning and imitation. In EGT, the accumulated returns of the game are associated with the fitness of an individual, such that the most successful individuals would reproduce more often, and their strategies spread in the populations. Interestingly, such evolutionary framework is mathematically equivalent to social learning, where individuals revise their strategy looking at and imitating the success and actions of others that show more fit than they [9]. As a result, strategies that do well spread in the population.

Yet, contrary to social learning, more sophisticated agents (such as humans) might instead imagine how a better outcome could have turned out, if they would have decided differently, and thence self-learn by revising their strategy. This is where Counterfactual Thinking (CT) comes in. In this chapter, we propose a simple mathematical model to study the impact on cooperation of having a population of agents resorting to such counterfactual kind of reasoning, when compared with a population of just social learners. Specifically, in this chapter, we pose three main questions:

1. How can we formalize counterfactual behavioural revision in large populations (taking cooperation dynamics as an application case study)?

2. Will cooperation emerge in collective dilemmas if, instead of evolutionary dynamics and social learning, individuals revise their choices through counterfactual thinking?

3. What is the impact on the overall levels of cooperation of having a fraction of counterfactual thinkers in a population of social learners? Does cooperation benefit from such diversity in learning methods?

To answer these questions, we develop a new population dynamics model based on evolutionary games, which allows for a direct comparison between the behavioural dynamics created by individuals who revise their behaviours through social learning and through counterfactual thinking. We consider individuals who interact in a public goods problem in which a threshold of players, less than the total group size, is necessary to produce benefits, with increasing contributions leading to increasing returns. This setup is common to many social dilemmas occurring in Nature and societies [15-17], combining a N-player interaction setting with non-linear returns. We show that such counterfactual players can have a profound impact on the levels of cooperation. Moreover, we show that just a small prevalence of counterfactual thinking within a population is sufficient to lead to higher overall levels cooperation when compared with a population made up solely of social learners.

To our knowledge, this is the first time that counterfactual thinking is considered in the context of the evolution of cooperation in populations of agents, by employing evolutionary games with both counterfactual thinking and social learning to that effect. Nevertheless, other works have made use of counterfactuals in the multi-agent context. Foerster et al. [18] addressed cases where counterfactual thinking consists in imagining changes in the rules of the game, an interesting problem not addressed in this chapter. Peysakhovich et al. [19] rely on the availability and use of a centralized critic that estimates counterfactual advantages for the multi-agent system by reinforcement learning policies. While interesting from a engineering perspective, it differs significantly from our work, in that we have no central critic, no utility function to be maximized, nor the aim of conjuring a policy that optimizes given utility function. To the contrary, in our approach the population evolves, leading to emergent behaviours, by relying on social learning and counterfactual thinking given at the start to some of the agents. Hence the two approaches deal with distinct problem settings and are not comparable. A similar approach is used by Colby et al. [20] where there is a collective utility function to be maximized. Finally, in Ref. [7], it is adopted a modelling framework that neither considers a population nor multi-agent cooperation, but individual counterfactual thinking about another agent's intention. It does this by counterfactually imagining whether another agent's goal would still be achieved if certain harmful side effects of its actions had not occurred. If so, those harmful side effects were not essential for the other achieving its goal, and hence its actions were not immoral. Otherwise, they were indeed immoral because the harmful side effects were intended because necessary to achieve the goal.

This chapter is organized as follows. In Section 2 we detail the principles underlying our counterfactual thinkers and the N-person collective dilemma used to illustrate the idea. In Section 3 we introduce the mathematical formalism associated with counterfactual and social learning dynamics. Importantly, this formalism is independent of the game chosen. In Section 4 we show the results for the impact on cooperation dynamics of counterfactual reasoning when compared with social learning, and discuss the influence of counterfactual thinkers in hybrid population of both social learners and counterfactual agents. Section 5 provides a discussion on the results obtained, also drawing some suggestions for future works.

## 2. Counterfactual thinking and evolutionary games

Counterfactual thinking (CT) can be exercised after knowing one's resulting payoff following a single playing step with a co-player. It employs the counterfactual thought: *Had I played differently, would I have obtained a better payoff than I did?* This information can be easily obtained by consulting the game's payoff matrix, assuming the co-player would have made the same play, that is, other things being equal. In the positive case, the CT player will learn to next adopt the alternative play strategy.

A more sophisticated CT would search for a counterfactual play that improves not just one's payoff, but one that also contemplates the co-player not being worse off, for fear the co-player will react negatively to one's opportunistic change of strategy. More sophisticated still, the new alternative strategy may be searched for taking into account that the co-player also possesses CT ability. And the co-player might too employ a Theory of Mind (ToM)-like CT up to some level. We examine here only the non-sophisticated case, a model for simple (egotistic) CT.

In Evolutionary Game Theory (EGT), a frequent standard form of learning is so-called Social Learning (SL). It basically consists in switching one's strategy by imitating the strategy of a more successful individual in the population, compared to one's success. CT instead can be envisaged as a form of strategy update learning akin to debugging, in the sense that: *if my actual play move was not conducive to a good accumulated payoff, then, after having known the co-player's move, I can imagine how I would have done better had I made a different strategy choice*.

When compared with SL, this type of reasoning is likely to have a minor impact in games of cooperation with a single Nash equilibrium (or a single evolutionary stable strategy, in the context of EGT) such as the Prisoner's dilemma or the Public Goods game mentioned above, where defection-dominance prevails. However, as we illustrate below, counterfactual thinking has the potential to have a strong impact in games of coordination, characterized by multiple Nash Equilibria: CT will allow for a meta-reasoning on which equilibria provide higher returns. This is particularly relevant since conflicts often described as public goods problems, or as a Prisoner's Dilemma game, can be also interpreted as coordination prob-lems, depending on how the actual game parameters are ranked and assessed [15].

As an example, consider the two-person Stag-Hunt (SH) game [15]. Players A and B decide to go hunt a stag, which needs to be done together for maximum possibility of success. But each might defect and go by himself hunt a hare instead, which is more assured, because independent of what the other does, though less rewarding. The dilemma is that each is not assured the other will in fact go hunt the stag in cooperation and is tempted by the option of playing it safe by going hunt hare. Concretely, one may assume that hunting hare has a payoff of 3, no matter what the other does; hunting stag with another has a payoff of 4; and hunting stag alone has a payoff of 0. Hence the two-person stag hunt expectance payoff matrix:

|        | **B** stag | **B** hare |
|--------|------------|------------|
| **A** stag | 4  4       | 0  3       |
| **A** hare | 3  0       | 3  3       |

A simple analysis of this payoff table would tell us that one should always adopt the choice of our opponent, i.e., to coordinate our actions, despite the fact that we may end in a sub-optimal *equilibria* (both going for hare).

The nature of this dilemma is generalizable to an N-player situation where a group of $N$ is required to hunt stag [17]. Let us then consider a group of $N$ individuals, who can be either cooperators (C) or defectors (D); the $k$ Cs in $N$ contribute a cost $c$ to the public good, whereas Ds refuse to do so. The accumulated contribution is multiplied by an enhancement factor $F$, and the ensuing result equally distributed among all individuals of the group, irrespective of whether they contributed or not. The requirement of coordination is introduced by noticing that often we find situations where a minimum number of Cs is required within a group to create any sort of collective benefit [17, 21]. From group hunting [17] and the rise and sustainability of Human organizations [22], to collective action to mitigate the effects of dangerous climate change [23], examples abound where a minimum number of contributions is required before any public good is produced.

Following [17], this can be modelled through the addition of a coordination threshold *M*, leading to the following straight forward payoff functions for the Ds and Cs — Cs contribute a cost c to the common pool and Ds refuse to do so — where *j* stands for the number of contributing *k*:

$$P_D(j) = H(j-M)\,jFc/N$$
$$P_C(j) = P_D(j) - c \tag{1}$$

respectively. Here, *H* represents the Heaviside step function, taking the value of $H(x)=1$ if $x \geq 0$, and $H(x)=0$ otherwise. Above the coordination threshold *M*, the accumulated contribution (*j.c*) is increased by an enhancement factor *F*, and the total amount is equally shared among all the *N* individuals of the group. This game is commonly referred to as the N-person Stag-Hunt game [17]. For *M*=1, one recovers the classical N-person Prisoner's dilemma and the Public Goods game alluded to in the introduction.

Here we will consider a population of agents facing this N-person Stag-hunt dilemma, revising their preferences through social learning and through counterfactual thinking. The following section provides the details of how the success of an agent is computed and how agents revise their strategies in each particular case. The mathematical details are given for the sake of completeness, but a detailed understanding of the equations is not required to follow the insights of the model. For more information on evolutionary game theory and evolutionary dynamics in finite population, we refer the interested reader to reference [9].

## 3. Population dynamics with social learning and counterfactual thinking

Let us consider finite population of *Z* interacting agents where all agents are equally likely to interact with each other. In this case, the success (or fitness) of an agent results from the average payoff obtained from randomly sampling groups of size *N<Z*. As a result, all individuals adopting a given strategy will share the same fitness. One can compute the average fitness of each individual through numerical simulations, averaging over the payoff received in a large number of groups randomly sampled from the population. Equivalently, one can also compute analytically the average fitness of a strategy assuming a random sampling of groups, and averaging over the returns obtained in each group configuration[1].

Using this framework as a baseline model for the interactions among agents, let us now detail how population evolution proceeds under social learning (SL) and under counterfactual thinking (CT). Our $Z$ interacting agents can either resort to SL or CT to revise their behaviours. If an agent $i$ resorts to SL, $i$ will imitate a randomly chosen individual $j$, with a probability $p$ that augments with the increase in fitness difference between $j$ and $i$, given by $f_i$ and $f_j$ respectively. This probability can take many forms. We adopt the ubiquitous standard Fermi distribution to define the probability $p$ for SL

$$p_{SL} = \left[1 + e^{-\beta_{SL}[f_j - f_i]}\right]^{-1} \qquad (2)$$

in which $\beta_{SL}$ expresses the unavoidable noise associated with errors in the imitation process [24, 25]. Hence, successful individuals might be imitated with some probability, and the associated strategy will spread in the population.

Given the above assumptions, it is easy to write down the probability to change the number $k$ of Cs (by ±1 at each time step) in a population of $Z$-$k$ Ds in the context of social learning. The number of Cs will increase if an individual D has the chance to meet a role model with the strategy C, which occurs with a probability $\dfrac{Z-k}{Z}\dfrac{k}{Z-1}$. Imitation will then effectively occur with a probability $\left[1+e^{-\beta_{SL}[f_C-f_D]}\right]^{-1}$ (see Eq. 2), where $f_C$ and $f_D$ are the fitness of a C and of a D, respectively. Similarly, to decrease the number of Cs, one needs a C meeting a random role model D, which occurs with a probability $\frac{k}{Z}\frac{Z-k-1}{Z-1}$; imitation will then occur with a probability given by Eq. 2. Altogether, one may consequently write the probability that one more or one less individual ($T^+$ and $T^-$ in Figure 1) adopts a cooperative strategy through social learning as[2]

$$T_{SL}^{\pm}(k) = \frac{k}{Z}\frac{Z-k}{Z}\left[1+e^{\mp\beta_{SL}[f_B(k)-f_A(k)]}\right]^{-1}. \qquad (3)$$

---

[1] Formally, one can write this process as an average over a hyper-geometric sampling in a population of size $Z$ and $k$ cooperators. This gives the probability of an agent to interact with $N$-1 other players, where, among those, $j$ are cooperators. In this case, the average fitness $f_D$ and $f_C$ of $D$s and $C$s, respectively, in a population with $k$ $C$s, is given by [17] $f_D(k) = \binom{Z-1}{N-1}^{-1} \sum_{j=0}^{N-1} \binom{k}{j}\binom{Z-k-1}{N-j-1} P_D(j)$ and $f_C(k) = \binom{Z-1}{N-1}^{-1} \sum_{j=0}^{N-1} \binom{k-1}{j}\binom{Z-k}{N-j-1} P_C(j+1)$, where $P_C$ and $P_D$ are given by Eq. (1).

Differently, agents that resort to CT assess alternatives to their present returns, had they used an alternative play choice *contrary* to what actually took place. Agents imagine how the outcome could have worked if their decision (or strategy, in this case) would have been different. In its simplest form, this can be modelled as an incipient form of myopic best response rule [13] at the population level, taking into account the fitness of the agent in a configuration that did not, but could have occurred. In the case of CT, an individual $i$ adopting a strategy $A$ will switch to $B$ with a probability

$$p_{CT} = \left[1+e^{-\beta_{CT}\left[f_i^B-f_i^A\right]}\right]^{-1} \qquad (4)$$

that increases with the fitness difference between the fitness the agent would have had if it had played $B$ ($f_i^B$), and the fitness the agent actually got by playing $A$ ($f_i^A$). This can easily be computed considering the fitness players of C or players of D would have had in a population having an additional cooperator, or an extra defector, depending on the alternative strategy chosen by the individual revising its strategy. As before, one may write down the probability to change the number $k$ of Cs by plus or minus 1. To increase the number of Cs, one would need to select a D — which occurs with a probability $(Z-k)/k$ — and this D counterfactually decides to switch to C with the probability given by Eq. (4):

---

[2] For simplicity we assume that the population is large enough such that $Z \approx Z-1$.

$$T_{CT}^{+}(k) = \frac{Z-k}{Z}\left[1+e^{-\beta_{CT}\left[f_C(k+1)-f_D(k)\right]}\right]^{-1} \quad (5)$$

Similarly, the probability to decrease the number of Cs by one through counterfactual thinking would be given by

$$T_{CT}^{-}(k) = \frac{k}{Z}\left[1+e^{-\beta_{CT}\left[f_D(k-1)-f_C(k)\right]}\right]^{-1} \quad (6)$$

This expression is slightly simpler than the transition probabilities of SL due to the fact that, in this case, the reasoning is purely individual, and does not depend on the existence and fitness of meeting individuals adopting a different strategy.

Importantly, CT assumes that agents get access to the returns under different actions. This is, of course, a strong assumption that precludes the use of CT in all levels of complexity and evolutionary settings. Nonetheless, we assume that this feature is shadowed by some sort of error through the parameter $\beta_{CT}$, which, once again, expresses the noise associated with guessing the fitness values.

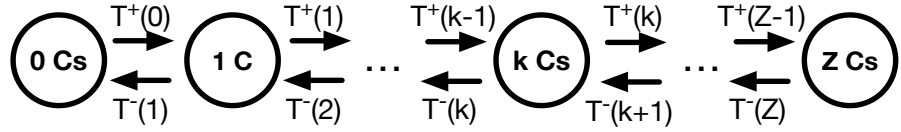

Fig 1. Social dynamics as a Markov chain. Representation of the transitions among the $Z+1$ available states, in a population of size $Z$, and two strategies ($C$ and $D$). Both social learning $T_{SL}^{\pm}(k)$ and counterfactual thinking $T_{CT}^{\pm}(k)$, where $k$ stands for the number of cooperators ($C$s), create a one-dimensional birth and death stochastic process of this type. One may also inquire about the stationary distribution $s_k$ for each case, which represent the expected fraction of time a population spends in each given state, in a large timespan.

We further assume that, with probability $\mu$, individuals may switch to a randomly chosen strategy, freely exploring the space of possible behaviours. Thus, with probability $\mu$, there occurs a mutation and an individual adopts a random strategy, without resorting to any of the above fitness dependent heuristics. With probability $(1-\mu)$ we have either social learning or counterfactual reasoning learning. As a result, for both SL and CT transition probabilities, we get modified transition probabilities given by

$$T^+_{SL/CT}(k,\mu) = (1-\mu)T^+_{SL/CT}(k) + \mu(Z-k)/Z \qquad (7)$$

for the probability to increase from *k* to *k+1* Cs and

$$T^-_{SL/CT}(k,\mu) = (1-\mu)T^-_{SL/CT}(k) + \mu k/Z \qquad (8)$$

for the probability to decrease to *k-1* (see Figure 1).

These transition probabilities can be used to assess the most probable direction of evolution in SL and CT. This is given by a learning gradient (often called gradient of selection [17, 23] in the case of SL), expressed by

$$G_{SL}(k) = T^+_{SL}(k,\mu) - T^-_{SL}(k,\mu) \qquad (9)$$

and

$$G_{CT}(k) = T^+_{CT}(k,\mu) - T^-_{CT}(k,\mu) \qquad (10)$$

respectively. When $G_{SL}(k) > 0$ and $G_{CT}(k) > 0$ ($G_{SL}(k) < 0$ and $G_{CT}(k) < 0$), time evolution is likely to act to increase (decrease) the number of Cs. In other words, for a given number *k* of Cs, the sign of *G(k)* offers the most likely direction of evolution.

This mathematical framework allows one to obtain the fraction of time that the population spends in each configuration after a long time has elapsed. To do so, it is important to note that the above transition probabilities define a stochastic process in which the probability of each event depends only on the current state of the population. In other words, we are facing a Markov process, whose states are given by the number of cooperators $k \in \{0,...,Z\}$. The transitions among all *Z+1* states can be seen as a transition matrix $\Lambda_{ij}$ such that $\Lambda_{k,k\pm 1} = T^{\pm}_{SL/CT}(k,\mu)$ and $\Lambda_{k,k} = 1 - \Lambda_{k,k-1} - \Lambda_{k,k+1}$. The average time the system spends in each state *k* is given by the so-called *stationary distribution* $s_k$, which is obtained from the eigenvector corresponding to the eigenvalue 1 of the transition matrix $\Lambda$ [26].

Finally, we can also use the stationary distribution to define a global *cooperation index*

$$\langle C \rangle = \sum_k k s_k \qquad (11)$$

which gives the number of Cs across states (*k*), weighted by the time the system spends in each state, i.e., by the stationary distribution $s_k$.

## 4. A comparison of social learning and counterfactual prompted evolutions, and an analysis of their interplay

In Figure 2a, we illustrate the behavioural dynamics both under CT and SL for the same parameters of the N-person Stag-hunt game. For each fraction of co-operators (Cs), if the gradient *G* (for both SL or CT) is positive (negative), then it is likely the fraction of Cs will increase (decrease). As shown, in both cases, the dynamics is characterized by two basins of attraction and two interior fixed points[3]: one unstable (also known as a coordination point), and a stable co-existence state between Cs and Ds. To achieve stable levels of cooperation (in a co-existence state), individuals must coordinate to be able to reach the cooperative basin of attraction on the right-hand side of the plot, a common feature in many non-linear public goods dilemmas [17, 23]. Figure 2a also shows that CT allows for the creation of new playing strategies, absent before in the population, since new strategies can appear spontaneously based on individual reasoning. By doing so, CT interestingly leads to different results if compared to SL. In this particular scenario, it is evident how CT may facilitate coordination of action, as individuals can reason on the sub-optimal outcome associated with non-reaching the coordination threshold, and individually react to that.

In Figure 2b, we show the stationary distribution of the Markov chain associated with the transition probabilities indicated above, showing how cooperation can benefit from CT. The stationary distribution characterizes the prevalence in time of each fraction of co-operators (*k/Z*). In this particular configuration, it is shown how in SL (black line), the population spends most of the time in low values for the fraction of co-operators. Whenever CT is allowed, cooperation is maintained most of the time. This emerges from the new position of the unstable fixed point shown in Figure

---

[3] Strictly speaking, by the finite population analogues of the internal fixed points in infinite populations.

2a. We further confirmed (not shown) the equivalence of CT and SL prompted evolutions, in the absence of coordination constraints (i.e., when *M*=1). In this case, we would have a defection dominance dilemma with a single attractor at 100% of defectors. Thus, in this regime, CT will have a marginal impact. Nonetheless, as long as the N-person game includes the need for coordination, translated in the existence of (at least) two basins of attraction and an internal unstable fixed point, CT may have the positive impact shown in Figure 2. In this particular case of the N-person Stag-Hunt dilemma, as shown by Pacheco et al. [17], the existence of these two basins of attraction depends on the interplay between *F*, M and the group size, *N*.

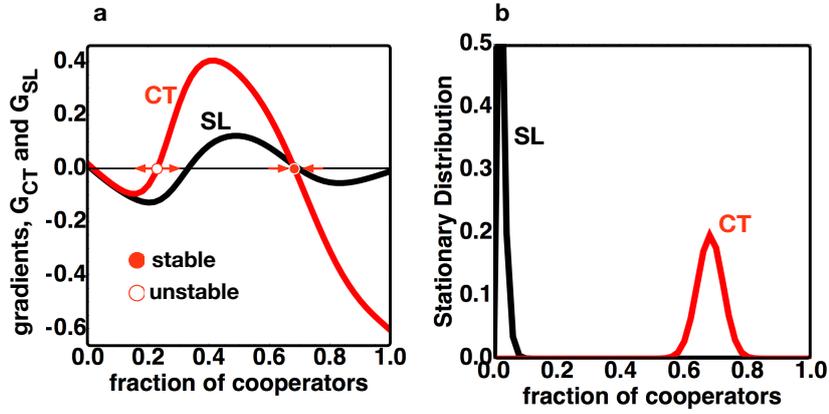

Fig. 2. Left panel: Learning gradients for social learning (black line) and counterfactual thinking (red line) — see $G_{SL}(k/Z)$ in see Eq. 9 and $G_{CT}(k/Z)$ in Eq. 10 — for the N-person SH game ($Z$=50, $N$=6, $F$=5.5. $M$=N/2, $c$=1.0, $\mu$=0.01, $\beta_{SL}=\beta_{CT}$=5.0). For each fraction of co-operators, if the gradient is positive then it is likely that the number of co-operators will increase; for negative gradient values cooperation is likely do decrease. Empty and full circles represent the finite population analogue of unstable and stable fixed points, respectively. Right panel: Stationary distribution of the Markov processes created by the transition probabilities pictured in the left panel; it characterizes the prevalence in time of each fraction of co-operators in finite populations (see main text). Given the positions of the roots of the gradients and the amplitudes of $G_{CT}$ and $G_{SL}$, contrary to the SL configuration, in the CT case the population spends most of the time in a co-existence between Cs and Ds.

Until now, individuals can either revise their strategies through social learning or counterfactual reasoning. However, one could also envisage situations where each agent may resort to CT and to SL in different circumstances, a situation prone to occur in Human populations. To encompass such heterogeneity at the level of agents, let us consider a simple model in which agents resort to SL with a probability χ, and to CT with a probability (1-χ), leading to a modified learning gradient given by $G(k) = \chi G_{SL}(k) + (1-\chi) G_{CT}(k)$.

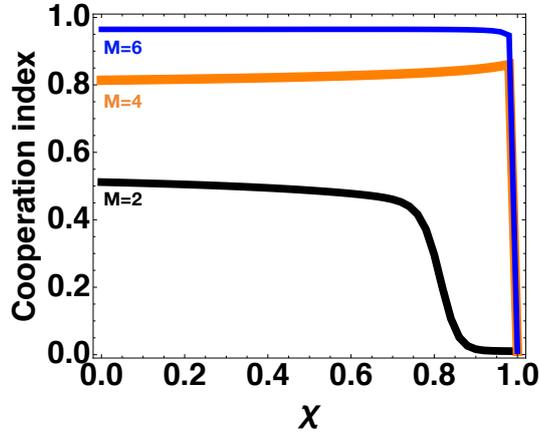

Fig. 3. Overall level of cooperation (cooperation index, see Eq. 11) as a function of the prevalence of individuals resorting to social learning (SL, χ) and counterfactual reasoning (CT, 1-χ). We show that only a relatively small prevalence of counterfactual thinking is required to nudge cooperation in an entire population of self-regarding agents. Other parameters: $Z=50$, $N=6$, $F=5.5$, $c=1.0$, $\mu=0.01$, $\beta_{SL}=\beta_{CT}=5.0$.

In Figure 3, we show the impact χ on the average cooperation levels (see Eq. 11) in a N-person Stag-Hunt dilemma in which, in the absence of CT, cooperation is unlikely to persist. Remarkably, our results suggest that a tiny prevalence of individuals resorting to CT is enough to nudge an entire population of social learners towards highly cooperative standards, providing further indications on the robustness of cooperation prompted by counterfactual reasoning.

## 5. Discussion

In this contribution we illustrate how decision-making shaped by counterfactual thinking (CT) is worth studying in the context of large populations of self-regarding agents. We propose a simple form to describe the population dynamics arising from CT and study its impact in behavioural dynamics when individuals face a threshold public goods problem. We show that CT enables the arise of pro-social behaviour in collective dilemmas, even where non or little existed before. We do so in the framework of non-cooperative N-person evolutionary games, showing how CT is able to modify the *equilibria* expected to occur when individuals revise their choices through social learning (SL). Specifically, we show that CT is particularly effective in easing the coordination of actions by displacing the unstable fixed points that characterize this type of dilemmas. In the absence of a clear need to coordinate (e.g., whenever $M=1$ in the N-person game discussed) CT offers equivalent results to those obtained with SL. Nonetheless, this is an especially gratifying result since many of the mechanisms known to promote cooperation in defection-dominance dilemmas (e.g., Prisoner's Dilemma, Public Goods games, etc.) enlarge the chances of cooperation by transforming the original dilemma into a coordination problem [10]. Thus, CT has the potential to be even more effective when applied in combination with other known mechanisms of cooperation, such as conditional strategies based on past actions, commitments, signalling, emotional reactions, or reputations based dynamics [16, 27-29]. Moreover, it is worth pointing out that the NPSH dilemma adopted here as a particular case study — which combines co-existence and coordination dynamics (see Figure 2a) — also represents the dynamics that emerge from most N-person threshold games, and from standard Public Goods dilemmas in the presence of group reciprocity [30], quorum sensing [16], and adaptive social networks [31], which further highlights the generality of the insights provided.

We also analyse the impact of having a mixed population of CT and SL, since it is unlikely that individuals would resort to a single heuristic for strategy revision. We show that, even when agents seldom resort to CT, highly cooperative standards are achieved. This result may have various interesting implications, if heterogeneous populations are considered. For instance, we can envision a near future made of hybrid societies compris-ing humans and machines [32-34]. In such scenarios, it is not only important to understand how human behaviour changes in the presence of artificial entities, but also to understand which properties should be included in artificial agents capable of leveraging cooperation in such hybrid collectives [32]. Our results suggest that a small fraction of artificial CT agents in a population of Humans social learners can decisively influence the dynamics of cooperation towards a cooperative state. These insights should be confirmed through a two-population ecological model where SLs influence CTs (and vice-versa), but also by including CTs that have access to a lengthier record of plays, rather than just the last one, or learn from past actions, creating a time-dependence that may be assessed through numerical computer simulations. Work along these lines is in progress.

***Acknowledgements.*** We are grateful to The Anh Han and Tom Lenaerts for comments. We are also grateful to the anonymous reviewers for their improvement recommendations This work was supported by FCT-Portugal/MEC, grants NOVA-LINCS UID/CEC/04516/2013, INESC-ID UID/CEC/50021/2013, PTDC/EEI-SII/5081/2014, and PTDC/MAT/STA/3358/2014.